\documentclass[sigconf]{aamas}  

\AtBeginDocument{%
  \providecommand\BibTeX{{%
    \normalfont B\kern-0.5em{\scshape i\kern-0.25em b}\kern-0.8em\TeX}}}
    
\usepackage{booktabs}    
\usepackage{flushend} 
\usepackage{amsmath, amsfonts}
\usepackage[linesnumbered,lined,boxed,commentsnumbered]{algorithm2e}
\usepackage{algorithmic}
\usepackage{subcaption}
\usepackage{hyperref}
\usepackage{enumitem}

\setcopyright{ifaamas}  
\copyrightyear{2020} 
\acmYear{2020} 
\acmDOI{} 
\acmPrice{} 
\acmISBN{} 
\acmConference[AAMAS'20]{Proc.\@ of the 19th International Conference on Autonomous Agents and Multiagent Systems (AAMAS 2020)}{May 9--13, 2020}{Auckland, New Zealand}{B.~An, N.~Yorke-Smith, A.~El~Fallah~Seghrouchni, G.~Sukthankar (eds.)}  

\begin{document}

\title{Learning Context-aware Task Reasoning for Efficient Meta-reinforcement Learning } 


\author{Haozhe Wang}

\email{wanghzh@shanghaitech.edu.cn}
\affiliation{%
 \institution{ShanghaiTech University}
 \city{Shanghai}
 \state{china}
}

\author{Jiale Zhou}

\email{zhoujl@shanghaitech.edu.cn}
\affiliation{%
 \institution{ShanghaiTech University}
 \city{Shanghai}
 \state{china}
}

\author{Xuming He}

\email{hexm@shanghaitech.edu.cn}
\affiliation{%
 \institution{ShanghaiTech University}
 \city{Shanghai}
 \state{china}
}










\begin{abstract}
Despite recent success of deep network-based Reinforcement Learning (RL), it remains elusive to achieve human-level efficiency in learning novel tasks. While previous efforts attempt to address this challenge using meta-learning strategies, they typically suffer from sampling inefficiency with on-policy RL algorithms or meta-overfitting with off-policy learning. In this work, we propose a novel meta-RL strategy to address those limitations. In particular, we decompose the meta-RL problem into three sub-tasks, task-exploration, task-inference and task-fulfillment, instantiated with two deep network agents and a task encoder. During meta-training, our method learns a task-conditioned actor network for task-fulfillment, an explorer network with a self-supervised reward shaping that encourages task-informative experiences in task-exploration, and a context-aware graph-based task encoder for task inference. 
We validate our approach with extensive experiments on several public benchmarks and the results show that our algorithm effectively performs exploration for task inference, improves sample efficiency during both training and testing, and mitigates the meta-overfitting problem.    
\end{abstract}

%


\maketitle

\newcommand{\defeq}{\overset{\text{\tiny def}}{=}}
\def\T{\mathcal{T}}
\def\E{\mathbb{E}}

\section{Introduction}
Modern reinforcement learning has achieved great successes in solving certain complex tasks by utilizing deep neural networks, which can even be trained from scratch~\cite{alphazero, poker, alphastar}. Such successes, 
however, require a large amount of training experiences for new tasks. In contrast, human 
learners are able to exploit past experiences when facing a novel problem and quickly learn 
skills for a related task~\cite{lake2017building}. To achieve such fast adaptation is a critical step 
towards building a general AI agent capable of solving multiple tasks in real-world environments. 

A principled way to tackling the problem of efficient adaptation is the meta 
learning framework~\cite{finn2017model}, which aims to capture shared knowledge across 
tasks and hence enables an agent to learn a similar task using only a few experiences. 
In the reinforcement learning setting, however, as the learning agent also needs to explore 
in each novel task, it is particularly challenging to design an efficient meta-RL algorithm.       
A majority of prior works adopt on-policy RL algorithms~\cite{rl-squared,finn2017model,gupta2018meta,rothfuss2018promp}, 
which are data-inefficient during 
meta-training~\cite{rakelly2019efficient}. 
To remedy this, \citet{rakelly2019efficient} propose an alternative strategy, PEARL, that 
relies on off-policy RL methods to achieve sample efficiency in meta-training. 
By introducing a latent variable to represent a task, their method decomposes the problem into online task inference and task-conditioned policy learning that uses experiences from a replay buffer (i.e., off-policy learning). 

Nevertheless, such an off-policy strategy has several limitations during meta-test stage, particularly for the few-shot setting. First, it ignores the role of exploration in the task inference (cf. meta-episodes in \cite{rl-squared}), which is critical in efficient adaptation as the exploration is responsible for collecting informative experiences within the few episodes to infer tasks. 
In addition, the agent has to explore in an online fashion for 
task inference during meta-test, which typically has a different sample distribution from the replay buffer that provides adaptation data at meta-train stage. PEARL partially alleviates this problem of train-test mismatch~\cite{vinyals2016matching} by adopting a replay buffer of recently collected data. Such an in-between strategy~\cite{rakelly2019efficient}, however, still leads to severe  ``meta-overfitting'', particularly for online task inference (cf. Sec~\ref{sec:exp}\&\ref{exp:und}).    
Furthermore, the adaptation data acquired by exploration are simply aggregated by a weighted average pooling for task inference. As the experience samples are not iid, such aggregation fails to capture their dependency relations, which is informative for task inference.

In this work, we propose a context-aware task reasoning strategy for meta-reinforcement learning to address the aforementioned limitations. 
Adopting a latent representation for tasks, we formulate the meta-learning as a POMDP, which learns an approximate posterior distribution of the latent task variable jointly with a task-dependent action policy that maximizes the expected return for each task. 
Our main focus is to develop an adaptive task inference strategy that is able to effectively map from few-shot experiences of a task to its representation without suffering from the meta-overfitting.  

To achieve this, we design a novel task inference network that consists of an exploration policy network and a structured task encoder shared by all tasks. Learning an exploration policy allows us to explore a task environment more efficiently and to introduce regularization to bridge the gap between meta-train and meta-test, leading to better generalization. The structured task encoder is built on a context-aware graph network, which is input permutation-invariant and size-agnostic in order to cope with variable number of exploration episodes. Our graph network encodes task-related dependency in experience data samples, capable of capturing complex task statistics from exploration to achieve more data-efficient task inference. 

To train our meta-learner, we develop a variational EM formulation that alternately optimizes the exploration 
policy network that collects experiences for a task, the context-aware graph network that performs 
task inference based on the collected task data, and an action policy network that aims to complete 
tasks towards maximum rewards given the inferred task information. 
Our meta-learning objective, formulated under the POMDP framework, is composed of two state-action value functions for the two respective policies. The state-action value for the exploration policy is guided by a 
reward function that automatically balances between the informativeness of the explored experience
and the stochasity
for randomized behavior to narrow the gap between online rollouts and offline buffers.
In essence, our method decomposes the meta-RL problem into three sub-tasks, task-exploration, task-inference and task-fulfillment, in which we learn two separate policies for different purposes and a task encoder for task inference.

We evaluate our meta-reinforcement learning framework on four benchmarks with a diverse task 
distributions, in which our approach outperforms prior works by improving 
testing efficiency with better asymptotic performance (up to 300\%) while effectively mitigating the train-test mismatch by a large margin (up to 75\%). 
Our contributions can be summarized as follows:
\begin{itemize}[leftmargin=4mm]
  \item We propose a sample-efficient meta-RL algorithm that achieves the state-of-the-art performances on multiple benchmarks with diverse task distributions. 
  \item We present a dual-agents design with a reward shaping strategy that explicitly optimizes for the ability for task exploration and mitigates meta-overfitting.
  \item We develop a context-aware graph network for task inference that models 
  the dependency relations between experience data in order to efficiently infer task posterior.
\end{itemize}
\section{Related work}
\subsection{Meta-reinforcement Learning}\label{sec:meta-rl}
Prior meta reinforcement learning methods can be categorized into the following three  
lines of work. 

The first line of work adopts a learning-to-learn strategy~\cite{finn2017model, stadie2018some,gupta2018meta,rothfuss2018promp,xu2018learning}. In particular, MAML~\cite{finn2017model} meta-learns a model initialization from which to adapt the parameters 
with policy gradient methods. To tackle issues in computing second-order derivatives for MAML, ProMP~\cite{rothfuss2018promp} 
further proposes a low-variance curvature estimator to perform 
proximal policy optimization that bounds the policy change
Recently, MAESN~\cite{gupta2018meta} improves MAML with more temporally coherent exploration by injecting structured noise
via a latent variable conditioned policy, and enables fast learning of exploration strategies.

The second category of approaches uses recurrent or recursive models to directly meta-learn a 
function that takes experiences as input and generates a policy for an agent~\cite{mishra2017simple,reiforcelearn,rl-squared}. 
Among them, $\text{RL}^2$~\cite{rl-squared} trains a recurrent agent with on-policy 
meta-episodes that comprise a sequence of exploration and task-fulfillment episodes, 
which aims to maximize the expected return of the task-fulfillment episode. 
Their method essentially learns to extract task information from the first few rounds of exploration, encoded in the latent states of its RNN, and complete the task in the final episode based on the known task information (latent states). 

The first two groups of work often learn a single policy to perform exploration for policy adaptation and task-fulfillment, which overloads the agent with two distinct objectives. By contrast, we learn two separate policies, one focusing on exploration for policy adaptation and the other for task-fulfillment.


In the third line of work, \citet{rakelly2019efficient} and \citet{taskinf} propose to first infer task with task experiences and adapts the agent according to the task knowledge. Framing meta-RL as a POMDP problem 
with probabilistic inference, \citet{taskinf} formulate task inference as belief 
state in POMDP, and learn the inference procedure in a supervised manner with privileged information 
as ground truth. 
PEARL \cite{rakelly2019efficient} learns to infer tasks in an unsupervised manner by introducing an extra  reconstruction term and incorporates off-policy learning to improve sample efficiency. 
However, 
both approaches ignore the role of exploration in task inference, and PEARL suffers from train-test mismatch in the data distribution for task inference.

By contrast, we propose to further disentangle meta-RL into task-exploration, task-inference and task-fulfillment, and introduce an exploration policy within a variational EM formulation. Our method enhances the inference procedure via active task exploration and effective task inference, achieves data-efficiency both in meta-train and meta-test, and mitigates meta-overfitting.

\subsection{Relational Modeling on Sets}

The problem of inference on a set of samples has been widely explored in literature~\cite{battaglia2018relational,gilmer2017neural,li2015gated, bruna2013spectral,kipf2016semi,hamilton2017inductive}, and here we focus on those approaches that learn a permutation-invariant and input size-agnostic model.
 
DeepSets~\cite{zaheer2017deep} proposes a general design principles for permutation invariant 
functions on sets. As an instantiation, Sets2Sets~\cite{vinyals2015order}
encodes a set using an attention mechanism, which retrieves vectors in a manner immune to the shuffle
of memory and accepts variable size of inputs. 
However, they typically do not consider relations among input elements, which is critical for modeling short experiences for task inference.
 
Self-attention module~\cite{internet,vain,santoro2017simple,wang2018non} is commonly used to model pairwise relationships between entities of a set.
Among them, Nonlocal Neural Networks~\cite{wang2018non} are capable of capturing global context using the scaled dot product attention \cite{vaswani2017attention}. Along this direction, 
graph networks~\cite{battaglia2018relational,gilmer2017neural} provide a flexible framework that can model arbitrary relationships among entities and is invariant to graph isomorphism. Self-attention on sets, 
a special case of graphs, can naturally be assimilated into the graph network framework~\cite{zhang2019latentgnn,battaglia2018relational,velivckovic2017graph}. 
Recently, LatentGNN~\cite{zhang2019latentgnn} develops a novel 
undirected graph model to encode non-local relationships through a shared latent space 
and admits efficient message passing due to its sparse connection. In this work, the design of our graph-based inference network is inspired by LatentGNN and DeepSets, aiming to efficiently model the dependency relationship
in experience data.

\section{Variational Meta-Reinforcement Learning}\label{sec:ps}
We aim to address the fast adaptation problem in reinforcement learning, which allows the learned agent to explore a new task (up to a budget) and adapts its policy to the new task. 
We adopt the meta-learning framework to enable fast learning, incorporating a prior of past experiences in structurally similar tasks. 

Formally, we assume a task distribution $P(T)$ from 
which multiple tasks are i.i.d sampled for meta-training and testing. 
Any task $T \sim P(T)$ can naturally be defined by its MDP $(S,A,\mu_0(s),\mathcal{T}, \gamma, r)$ in RL, where $S$ is the state space, $A$ is the action space, $\mu_0(s)$ is the distribution of the initial states, $\mathcal{T}(s^\prime|s,a)$ is the transition distribution, $\gamma$ is the discount in rewards which we will omit for clarity, and $r(s,a)$ is the reward function. 
Here the task distribution is typically defined by the distribution of the 
reward function and the transition function, which can vary due to some underlying parameters. 
We introduce a latent variable $z\sim P(Z)$ to represent the cause of task variations, and formulate the meta RL problem as a POMDP in which $z$ serves as the unobserved part of the state space. 



We first introduce two policies: an action policy $\pi$ that aims to maximize expected reward under a given task $z$, and an exploration policy $\xi$ for generating task-informative samples by interacting with the environment. We then adopt the off-policy Actor-Critic framework~\cite{dpg}, and learn two Q-functions $Q^\pi,Q^\xi$ for the action and exploration policy respectively. 
To achieve this, we define the following learning objective
\begin{equation}
	\begin{aligned}
		&\min_{Q^\pi,Q^\xi}\underset{
			\substack{\beta(s,a)\\
				\T(s^\prime,r|s,a)}
		}{\mathbb{E}}\left[\left(Q^\pi(s,a)-y\right)^2\right]\\
		&+\underset{
			\substack{\rho(s,a)\\
				\T(s^\prime|s,a),r^\xi
			}
		}{\mathbb{E}}\left[\left(Q^\xi(s,a)-y^\xi\right)^2\right]\label{eq:obj}
	\end{aligned}
\end{equation}

where $\beta,\rho$ denotes the $(s,a)$ distribution visited by the two behavior policies for $\pi,\xi$, $r^\xi$ is the reward for $\xi$ (Sec.~\ref{sec:exp}), $y\defeq r+\gamma V(s^\prime)$ is the Bellman Backup for $\pi$, and $y^\xi$ for $\xi$ defined similarly. Typically, $\beta,\rho$ is approximated by the experience replay buffers of $\pi,\xi$, denoted as $\mathcal{B},\mathcal{X}$ respectively~\cite{mnih2013playing,ddpg,haarnoja2018soft}.

The above Bellman Residual minimization can be interpreted as maximizing the log likelihood of the transition tuple $(s,a,s^\prime, r)$, i.e., 
\begin{align}
    \underset{
        \substack{\beta(s,a)\\
        \T(s^\prime,r|s,a)}
    }{\mathbb{E}}\left[\log P(s,a,y)\right].
    \label{eq:obj2}
\end{align}

To approach the approximate inference of the latent task variable $z$, we substitute Eq.~\eqref{eq:obj2} with the Q-learning objective for $\pi$ in Eq.~\eqref{eq:obj}\footnote{Although the Q-learning objective for $\xi$ has similar maximum likelihood interpretation, it is not included in the E-step for the approximate inference of the latent task variable $z$, because $z$ represents the tasks, but $r^\xi$ is not given by the environment. }, and utilize the variational EM learning strategy~\cite{bishop2006pattern} to maximize the objective. The latent task variable $z$ is seen as the unobserved state in a POMDP, and is often obtained through the approximate inference given past experiences~\cite{taskinf,rakelly2019efficient}. The past experience $x_t$ of time step $t$ is a list of transition tuples denoted as $x_t=\{(s_i,a_i,r_i, s_{i+1})\}_{i=0}^t$. we introduce a variational distribution $q(z|x)$ to approximate the intractable posterior $P(z|x)$, in which we alternate between optimizing for $q(z|x)$ and for other parameterized functions $Q^\pi, Q^\xi$
\footnote{The policies $\pi,\xi$ are also parameterized, but is learned separately with the guidance from their corresponding state-action value functions~\cite{dpg,ddpg,soft}.}.
We refer to $q(z|x)$ as the task encoder in our model.

For E-step, we maximize a variational lower bound $\mathcal{J}_e(q)$ derived from \eqref{eq:obj} to find the optimal $q(z|x)$. 

\begin{align}
    &\underset{
      \substack{(s,a,s^\prime,r)\sim \mathcal{B},\\
      x\sim \mathcal{X}, \\
                z\sim q(z|x)
              }
    }{\mathbb{E}}
    \left[-\left(Q^\pi(s,a,z)-y\right)^2-D_\text{KL}(q(z|x)\|P(z))\right]   \label{eq:obje}                   
\end{align}
Here replay buffers $\mathcal{B},\mathcal{X}$ overrides the previous notations of behavior policies $\beta(\cdot),\rho(\cdot)$. Note that the past experiences $x$ is sampled from the buffer $\mathcal{X}$ for task exploration policy $\xi$. $Q^\pi(s,a)$ in Eq.~\eqref{eq:obj} becomes $Q^\pi(s,a,z)$ in Eq.~\eqref{eq:obje} due to the expanded integral. The KL term serves as regularization, and $P(z)$ is initialized with the unit Gaussian prior $\mathcal{N(0,1)}$ at the beginning of a trajectory, and set to the previous posterior $q(z|x_{t-1})$ at time step $t$. 

For M-step, given $q(z|x)$, we minimize the following free-energy objective $\mathcal{J}_m(Q^\pi, Q^\xi)$ derived from \eqref{eq:obj} :
\begin{align}
  \underset{\substack{
    x\in\mathcal{X}\\
    z\sim q(z|x)\\
    (s,a,s^\prime)\sim\mathcal{B}
  }}{\mathbb{E}}\left[\left(Q^\pi(s,a,z)-y\right)^2\right]
  +
  \underset{\substack{
    x\in\mathcal{X}\\
    z\sim q(z|x)\\
    (s,a,s^\prime)\sim\mathcal{X}
  }}{\mathbb{E}}
  \left[\left(Q^\xi(s,a,z)-y^\xi\right)^2\right]
\end{align}
Based on this variational EM algorithm, our approach learns dual agents for task-exploration and task-execution respectively. The meta-training process interleaves the data collection process with the alternating EM optimization process, as shown in Alg.~\ref{algo}. 

\begin{figure}[t]
	\includegraphics[width=\columnwidth]{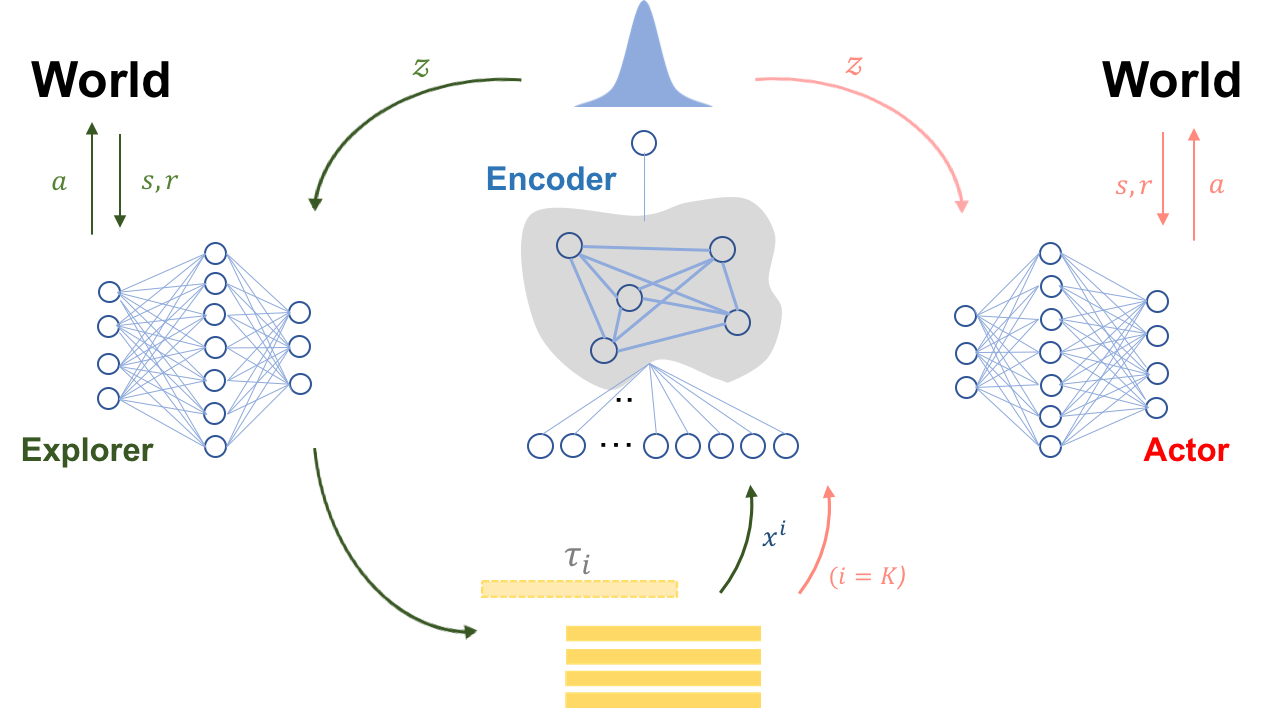}
	\caption{\small \textbf{Context-aware task reasoning for RL adaptation}. We separate the task into 
		task-exploration, task inference and task-fulfillment. The explorer interacts with the environment to collect experiences for the task encoder to update the belif of task. After an iterative process of $K-1$ rounds, the task encoder takes all the collections and gives the final task hypothesis to adapt the actor.}\label{fig:overview}
\end{figure}
\IncMargin{1em}
\begin{algorithm}[t]
	\SetAlgoLined
	\SetKwInOut{Input}{input}\SetKwInOut{Output}{output}
	\Input{Meta-train tasks $T_{train}=\{T_i\}_{1:M}\sim p(T)$, training steps $N$, number of E-step iterations $n_e$, number of M-step iterations $n_m$, number of tasks in a batch $n_b$, number of trajectories into task buffers $n_{trj}$, learning rate $\lambda$.}
	\Output{ $q(z|x),\pi,\xi,Q^\pi, Q^\xi$.}
	Initialize $q,\pi,\xi,Q^\pi, Q^\xi$, the dual variable $\alpha$, buffer $\mathcal{B}$ for $\pi$, buffer $\mathcal{X}$ for $\xi$.\\
	\While {not converged}{
		\For {task $T_i \in T_{train}$}{
			Collect $n_{trj}$ trajectories into buffers $\mathcal{B},\mathcal{X}$ with the policies $\pi,\xi$ respectively.
		}
		\For {each step $i$ in $N$ training steps}{
			randomly sample $n_b$ tasks from $T_{train}$ as set $T_{train}^B$.\\

			\eIf{
				$\textrm{i} \mod (n_e+n_m)<e$
			}{
				
				Get the inputs for the task set $T_{train}^B$ (E-step):\\ 
				$\left\{x_i\right\}_{i=1}^{n_b}\sim\mathcal{X}, \left\{(s_i,a_i,s_{i+1},r_i)\right\}_{i=1}^{n_b}\sim\mathcal{B}$.\\
				Update $q$ by SGD with loss $-\mathcal{J}_e(q)$.
			}{
				Get the inputs for the task set $T_{train}^B$ (M-step):
				$\left\{x_i\right\}_{i=1}^{n_b}\sim\mathcal{X}, \left\{(s_i,a_i,s_{i+1},r_i)\right\}_{i=1}^{n_b}\sim\mathcal{B},$\\
				$\left\{(s_i,a_i,s_{i+1},r_i)\right\}_{i=1}^{n_b}\sim\mathcal{X}$.\\
				Update $Q^\pi,Q^\xi$ by SGD with loss $\mathcal{J}_m(Q^\pi,Q^\xi)$.\\
				Update $\pi,\xi$ with $Q^\pi,Q^\xi$, e.g., SAC~\cite{haarnoja2018soft}.\\
				Update the dual variable $\alpha$ following \cite{soft}.
			}
			
		}
	}
	\caption{The Meta-training algorithm}\label{algo}
\end{algorithm} 
In contrast to \cite{taskinf,rakelly2019efficient}, our variational EM formulation derives from a unified learning objective and enables fast learning of dual agents with different roles. The explorer learns the ability of \emph{task-exploration} that targets at efficient exploration to (actively) collect sufficient task information for task inference,
while the actor learns for \emph{task-execution} that accomplishes the task towards high 
rewards. Both the explorer and the actor are task-conditioned, so that they are able to perform temporally-coherent exploration for different goals, given the current task 
hypothesis \cite{gupta2018meta}. 

With such a formulation, we need to answer the following questions to achieve
efficient and effective task inference to solve the meta-RL problem:
\begin{itemize}[leftmargin=4mm]
  \item How to guide the explorer towards active exploration for sufficient task information that is crucial in task inference with few experiences?
  \item How to achieve effective task inference that captures the relationships between experiences and reduce task uncertainty?
  \item Since we incorporate off-policy learning algorithms, how to mitigate the train-test mismatch issue due to the use of replay buffer 
   v.s. online rollouts during testing? 
\end{itemize}
We will address the above three questions in Sec.~\ref{sec:app}, which introduces our task inference strategy in detail.

\paragraph{Meta-test} At the meta-test time, we apply the same data collection procedure as in the meta-training stage.
For each task, the explorer $\xi(a|s,z)$ first samples a hypothesis $z$ from the posterior (initialized as $\mathcal{N}(0,1)$), and then explores optimally according to the hypothesis. The collected experiences $x$ during the exploration are used by the task encoder $q(z|x)$ to update the posterior of $z$. This process iterates until the explorer uses up the maximum number of rollouts. All the experiences are fed to the task encoder to produce a final posterior representing the belief of MDPs. We then sample from the posterior to generate a task-conditioned action policy $\pi(a|s,z)$, which interacts with the environment in an attempt to fulfill 
the task. This process is shown in Fig.~\ref{fig:overview}.

\section{Task Inference Strategy}\label{sec:app}

We now introduce our task reasoning strategy for meta-RL with limited data. Our goal is to collect experiences with sufficient information for task inference under limited budget, and to effectively infer the task posterior from the given task experiences. In Sec.~\ref{sec:exp}, we explain how we learn an explorer that pursues task-informative experiences with randomized behavior and mitigates train-test mismatch. In Sec.~\ref{sec:inf}, we elaborate on the network design of the task encoder, which enables relational modeling between task experiences.

\subsection{Learning the Exploration Policy}\label{sec:exp}

Our exploration policy aims to enrich the task-related information of experiences within limited trajectories. This goal is different from accomplishing the task and hence should not directly use the environment feedback. To this end, we design a reward function that jointly considers the quality and the diversity of the collected experiences. In the following section, we introduce the critical derivations of our reward function, and more details are put to the supplementary due to limited space. 

\subsubsection{Improving Quality}
The quality of a collected experience $c_t=(s_t,a_t,s_{t+1},r_t)$ can be quantified by comparing the change between two posteriors, $q(Z|X=x_{t})$ with the new experience and $q(Z|X=x_{t-1})$ without $c_t$. 
We adopt the KL divergence measure, and find that is has a convenient analytic form related with the Bellman Residual. The KL divergence computes the distance between the two posteriors:
\begin{equation}
  \begin{aligned}
    &\mathcal{D}_{KL}\left(q(Z|X=x_{t-1})\|q(Z|X=x_{t})\right) \\
    &=\underset{z\sim q(Z|X=x_{t-1})}{\E}\left[-\log q(Z=z|X=x_t)\right]+\text{constant}
  \end{aligned}\label{mutual}
\end{equation}
where $x_t = x_{t-1}\cup c_t$. The first term is the cross entropy, and the second term entropy is constant w.r.t. time step $t$. 
With Bayes Rule applied, we further expand the cross entropy term as:

\begin{equation}
  \begin{aligned}
    &\underset{\substack{z_{t-1}\sim q\left(\cdot|x_{t-1}\right)}}{\E}\left[-\log~q\left(Z=z_{t-1}|X=x_t\right) \right]\\
  &\propto \underset{\substack{z_{t-1}\sim q\left(\cdot|x_{t-1}\right)}}{\E}\left[ -\log \left(P(y_t|z_{t-1},s_t,a_t)\cdot \xi(a_t|s_t,z_{t-1})\right)\right]\\
  &\approx \sum_{z_{t-1}^{(i)}} \lambda \left\|Q^\pi\left(s_t,a_t,z_{t-1}^{(i)}\right)-y_t\right\|^2-\log \xi(a_t|s_t,z_{t-1}^{(i)})\label{curiosity}
  \defeq \hat{r}^\xi_t
  \end{aligned}
\end{equation}
  
where $z_{t-1}^{(i)}$ is the $i-$th Monte Carlo sample from  $q\left(\cdot|x_{t-1}\right)$ and the hyperparameter $\lambda$ controls reward scaling. 

The Bellman Residual term in Eq.~\eqref{curiosity} , is seen as a type of model-free reconstruction error for the transition tuple $c_t$ using previous set of task hypotheses $z_{t-1}$. 
Such reconstruction error can also be seen as a variant of intrinsic curiosity-based reward design~\cite{ICM}, which provides a self-supervised source of guidance from the execution policy $\pi$. 
The second term is the negative log-likelihood of the integral policy $\xi(a|s_t) = \int_{z} q(z|x_{t-1}) \xi(a|s_t,z)$ based on previous task hypotheses $z_{t-1}$, crediting higher for less likely actions. It also relates with the policy entropy as discussed in the next section.

\subsubsection{Improving Diversity}
To improve diversity, a direct solution is to add stochasity to the agent's behavior. We achieve this by adding a policy entropy constraint.
To this end, we formulate a constrained RL objective\footnote{Here we use the generic form of RL objective, but the discussion applies to the RL objective of other variants W.L.O.G., e.g., Q-learning. } for the exploration policy, similar to the entropy-regularized RL objective~\cite{soft}. 
\begin{equation}
  \begin{aligned}
    \E_{\rho(s,a,r)}\left[\sum_{i=0}^T r_i^\xi \right]\quad \textrm{s.t.} \quad \mathcal{H}\left(\xi(\cdot|s_t,z_{t-1})\right)\ge K
  \end{aligned}
\end{equation}
where $K$ is a hyperparameter for policy entropy threshold. \citet{soft} shows that such constrained objective can be treated as using an entropy-augmented reward function:
\begin{equation}
  \begin{aligned}
    &r^\xi_t \defeq \hat{r}^\xi_t + \alpha \mathcal{H}\left(\xi(\cdot|s_t,z_{t-1})\right) \\
    \approx &~\lambda \left\|Q^\pi\left(s_t,a_t,z_{t-1}^{(i)}\right)-y_t\right\|^2 -\alpha \log \xi(a_t|s_t,z_{t-1})
  \end{aligned}\label{ssr}
\end{equation}
where $\alpha$ is the dual variable. Monte Carlo Estimation is used to approximate the expectation in the policy entropy, which assimilates the negative log-likelihood term of Eq.~\eqref{curiosity}. Note that the Variational EM learning remains unchanged with this reward function, but with an additional SGD update for the dual variable $\alpha$~\cite{soft}. 

The reward function balances the informativeness and the diversity of an explorative action $a_t$ in state $s_t$, with an automatically tuned trade-off factor $\alpha$. The informativeness introduces supervision from the Bellman Residual of the execution policy $\pi$ as discussed in the previous section. The diversity is revealed in the policy entropy term, which promotes the randomized behavior. Such stochasity helps mitigate train-test mismatch, since the the online rollout data distribution is brought closer to the off-policy buffer data.




\subsection{Context-aware Task Encoder}\label{sec:inf}

Our task inference network computes the posterior of latent variable $z$ given a set of experience data, aiming to extract task information from experiences. To this end, we design a network module with the following properties:
\begin{enumerate}[leftmargin=4mm]
  \item \textit{Permutation-invariant}, as the output should not vary with the order of the inputs.
  \item \textit{Input size-agnostic}, as the network would encounter variable size of inputs within the arbitrary number of rollouts.
  \item \textit{Context-aware}, as extracting cues from a single sample should incorporate the context formed by other samples\footnote{Imagine in a 2d-navigation task where the agent aims to navigate to a goal location, a sample may indicate the possible location of the goal due to the high rewards, and can further eliminate possibilities by another sample that shows what locations are not possible by its low rewards.}. 
\end{enumerate}

\begin{figure}[t]
  \centering
  \includegraphics[width=\columnwidth]{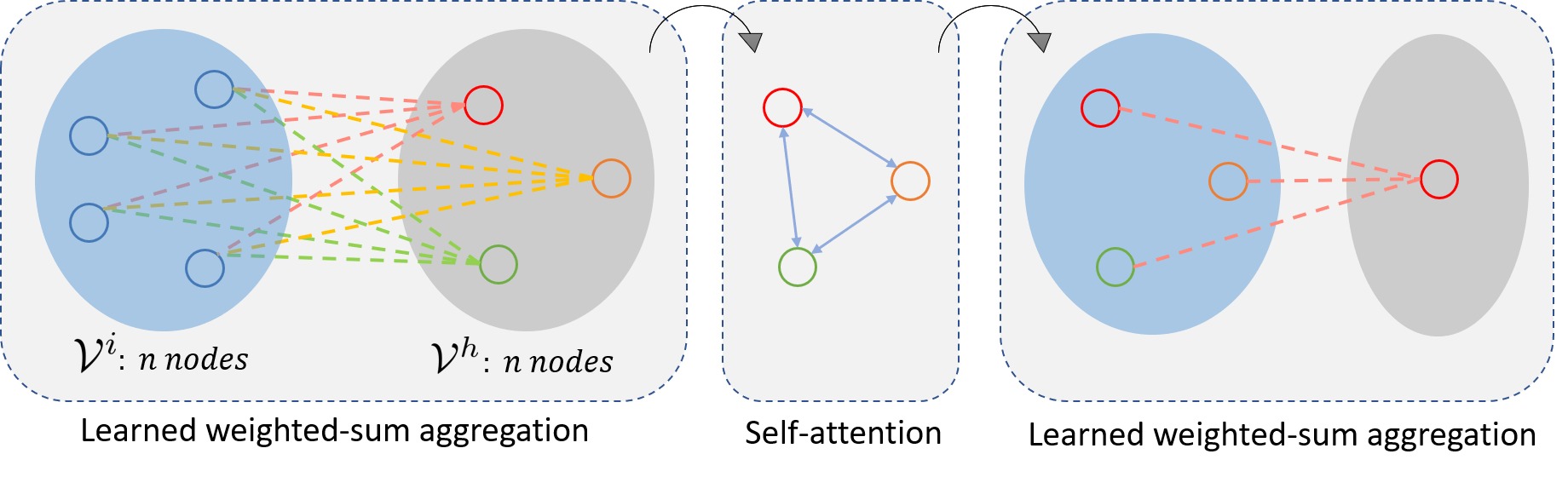}
  \caption{\small The task encoder.The first aggregation constructs a bipartite graph with full connections from the $n$ nodes in $V^i$ to the $c$ latent nodes in $V^h$. Self-attention operates on $V^h$, which are assembled to one latent node in the second aggregation.}
  
  \label{fig:inference}
\end{figure}

Specifically, we adopt a latent Graph Neural Network architecture~\cite{zhang2019latentgnn}, which integrates self-attention with learned weighted-sum aggregation layers. 
Formally, we introduce a set of latent node features $H=[h_1,\cdots,h_d]^T$ 
where $h_i\in \mathbb{R}^c$ is a c-channel feature vector same as the 
set of input node features $X=[x_i,\cdots,x_n]^T$. Note that $n$ can be 
arbitrary number while $d$ should be a fixed hyperparameter. We construct a graph 
$\mathcal{G}=(\mathcal{V},\mathcal{E})$ with the $n+d$ nodes and full connections 
from all of the input nodes to each of the latent nodes, i.e., 
$(v_i,v_j)
\in \mathcal{E}, v_i\in\mathcal{V}^i, v_j\in\mathcal{V}^h$ where 
$\mathcal{V}^i,\mathcal{V}^h$ refers to the set of input nodes and latent nodes respectively. The graph network module is illustrated in Fig.~\ref{fig:inference}.

The output of the aggregation layer $f_{\text{AGG}}$ is the latent node features $H$, which are computed as follows:
\begin{align*}
  h_k = \sum_{j=1}^n \psi(x_j,\phi_k)x_j, 1\leq k\leq d
\end{align*}
where $\psi(x,\phi_k)$ is a learned affinity function parameterized by $\phi_k$
that encodes the affinity between any input node $x$ and the $k^{th}$ latent node.
In practice, we instantiate this function as the dot product followed with normalization, i.e.,
$\text{softmax}_j (\phi_k x_j^T)$.

We then combine the above aggregation layer 
with the following self-attention layer $f_\text{ATN}$:
\begin{align*}
  \tilde{x}_j = \sum_{i=1}^n f(x_i,x_j)x_j, 1\leq i\leq n
\end{align*}
where we use the the scaled dot product attention \cite{vaswani2017attention}.

Following \cite{zhang2019latentgnn}, we propagate messages through a shared space 
with full-connections between latent nodes. We first pass the input nodes through an aggregation layer 
with $d$ latent nodes, where $d<<n$, then perform self-attention on the latent nodes (for multiple iterations), and finally 
pass the latent nodes through another aggregation layer with $1$ final latent node to obtain 
the final output, i.e., $f_\text{AGG}\circ f_\text{ATN}\circ f_\text{AGG}:
\mathbb{R}^{n\times c}\mapsto \mathbb{R}^{d\times c}\mapsto \mathbb{R}^{d\times c}\mapsto \mathbb{R}^{1\times c}$. The final output provides the parameters for $q(z|x)$ which is a Gaussian distribution. 
The network can be viewed as a multi-stage summarization process, in which we group the inputs 
into several summaries, and operate on these summaries to compute the relationships between entities, 
and produce a final summary of the entities and relationships.

\begin{figure*}[h]
  \centering
  \includegraphics[width=\textwidth]{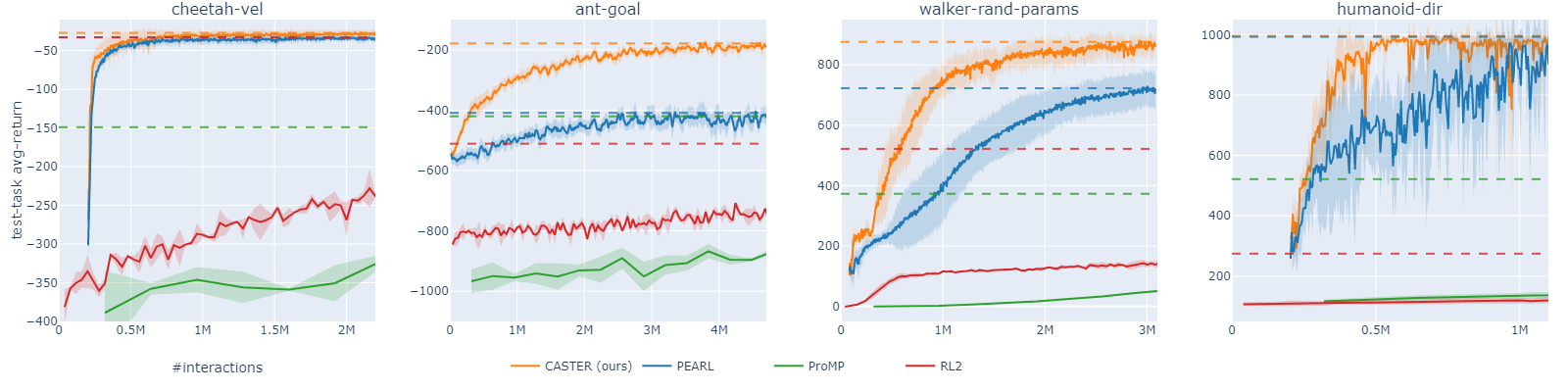}
  \caption{\small\textbf{Training efficiency.} The test-task performance versus the number of interactions with the environment during meta-training. The dash lines represent the asymptotic performance of each method. 
  }
  \label{fig:sota}
\end{figure*}
\begin{figure*}[h]
  \centering
  \includegraphics[width=\textwidth]{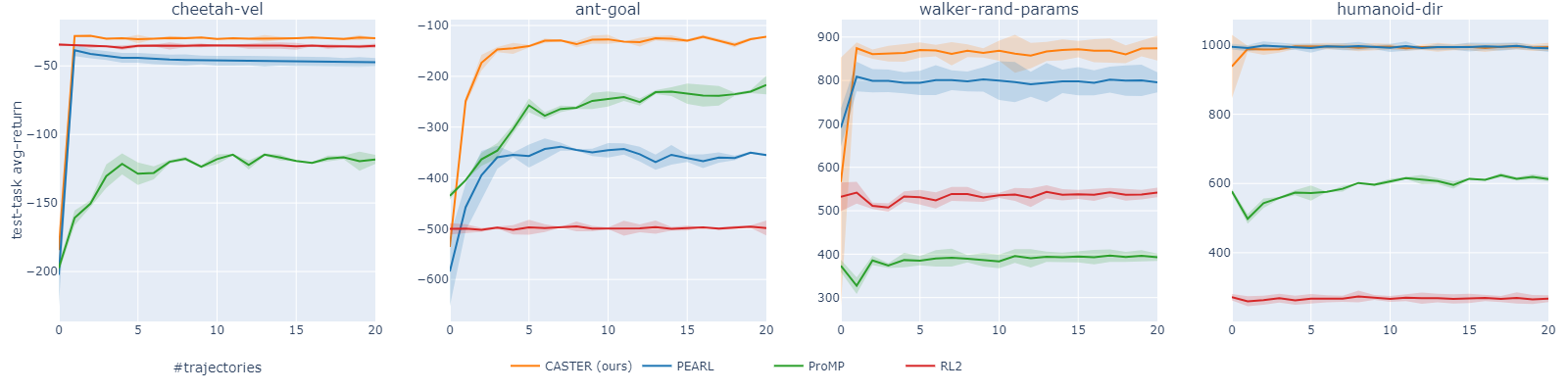}
  \caption{\small\textbf{Testing efficiency.} The x-axis denotes the number of trajectories used as adaptation data.}
  \label{fig:test-eff}
\end{figure*}
\section{Experiments}
In this section, we demonstrate the efficacy of our design and 
the behavior of our method, termed CASTER (shorthand for \emph{Context-Aware task inference with Self-supervised Task ExploRation for efficient meta-reinforcement learning}), through a series of experiments.
We first introduce our experimental setup in 
Sec.~\ref{exp:setup}. In Sec.~\ref{exp:sota}, we evaluate our CASTER against three meta-RL algorithms in terms of sample efficiency. We then compare the behavior of CASTER with PEARL 
regarding overfitting and exploration in Sec.~\ref{exp:und}. Finally, in Sec.~\ref{exp:abl}, we conduct ablation study on our design of the exploration strategy and the task encoder.

\subsection{Experiemtal Setup}\label{exp:setup}
We evaluate our method on four benchmarks proposed 
in \cite{rothfuss2018promp}\footnote{Two experiments ``cheetah-forward-backward'' and ``ant-forward-backward'' are not used because they only include two tasks (the goal of going forward and backward), and do not match the meta-learning assumption that there is a task distribution from which we sample the meta-train task set and a held-out meta-test task set. Such benchmark does not provide convincing evidence for the efficacy of meta-learning algorithms.} and an environment introduced in \cite{rakelly2019efficient} for ablation. They are implemented on the OpenAI Gym \cite{brockman2016openai} with 
the MuJoCo simulator \cite{todorov2012mujoco}. All of the experiments characterize locomotion tasks, 
which may vary either in the reward function or the transition function. We briefly describe the environments as follows.
\begin{itemize}[leftmargin=4mm]
  \item \textbf{Half-cheetah-velocity.} Each task requires the agent to reach a different target velocity.
  \item \textbf{Ant-goal.} Each task requires the agent to reach a goal location on a 2D plane .
  \item \textbf{Humanoid-direction.} Each task requires the agent to keep high velocity without falling off in a
  specified direction.
  \item \textbf{Walker-random-params.} Each task requires the agent to keep high velocity without falling
  off in different system configurations.
  \item \textbf{Point-robot.} Each task requires a point-mass robot to navigate to a different goal location on a 2D plane.
\end{itemize}
We adopt the following evaluation protocol throughout this section: first, the estimated per-episode performance on each task is averaged over (at least) three trials with different random seeds; second, the test-task performance is evaluated on the held-out meta-test tasks and is an average of the estimated
per-episode performance over all tasks with at least three trials.

\subsection{Performance}\label{exp:sota}
In this section, we compare with three meta-RL algorithms that are the representatives 
of the three lines of works mentioned in Sec.~\ref{sec:meta-rl}: 1) inference-based method, PEARL~\cite{rakelly2019efficient}; 2) optimization-based method, ProMP~\cite{rothfuss2018promp}; 3) black-box method, $\text{RL}^2$~\cite{rl-squared}.

For those baselines, we reproduce the experimental results via their officially released code, following their proposed meta-training 
and testing pipelines. 
We note that prior works~\cite{rl-squared,rothfuss2018promp} are not designed to optimize for sample efficiency but we keep their default hyperparameter settings in order to reproduce their results. 


\begin{figure*}[th]
	\centering
	\includegraphics[width=\linewidth]{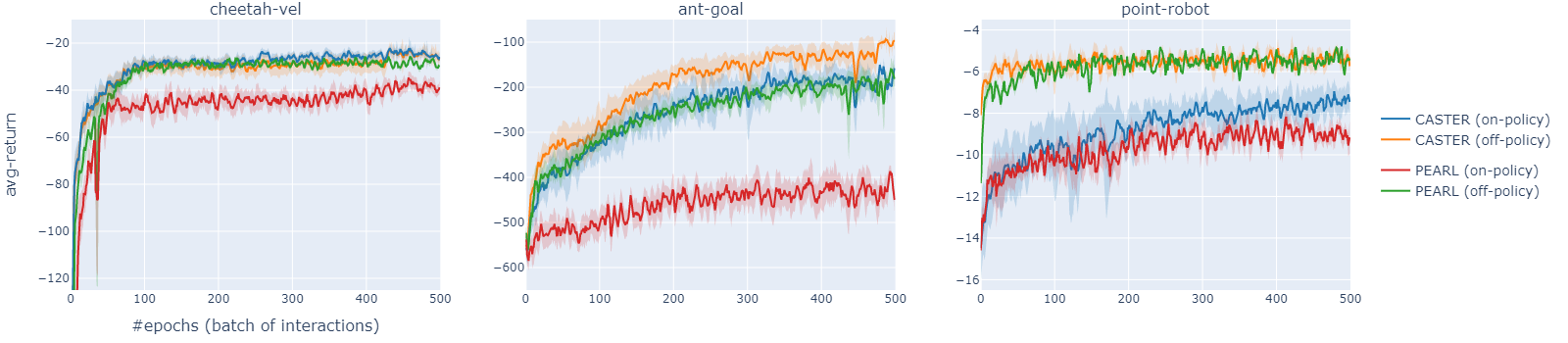}
	\caption{\small\textbf{Overfitting in off-policy meta-RL.} Each column in the plot corresponds to a different environment. We pick three environments most prone to meta-overfitting, i.e., `cheetah-vel', `ant-goal' and `point-robot' from left to right. }\vspace{-3mm}
	\label{fig:meta-over}
\end{figure*}

To demonstrate the training efficiency and testing efficiency, we plot the test-task performance as a function of the number of samples. Fig.~\ref{fig:sota} shows the comparison results on the training efficiency. Here the x-axis indicates the number of interactions 
with the environment used to collect buffer data. At each x-tick, we evaluate the test-task performance with \textbf{2} episodes for all methods. 
While for testing efficiency, the x-axis refers to the number of adaptation 
episodes used by different methods to perform policy adaptation (or task inference). 

We can see CASTER outperforms other methods by a sizable margin: CASTER achieves the same status of performance with much fewer environmental interactions while being able to reach higher performance. 
PEARL and CASTER incorporate off-policy learning and naturally enjoy an advantage in training sample efficiency. 
Our CASTER achieves better performances due to two reasons. On the one hand, efficient exploration potentially offers CASTER richer information within limited episodes (two in Fig.~\ref{fig:sota}), and pushes higher the upper bound of accurate task inference. On the other hand, context-aware relational reasoning extracts information effectively from the task experiences, and thus improves the ability of task inference.

Testing efficiency is shown in Fig.~\ref{fig:test-eff}, and CASTER still stands out against other methods. 
Notably, CASTER achieves large performance boost with the first several episodes, indicating that the learned exploration policy is critical in task inference. By contrast, other models either fail to improve from zero-shot performance (e.g. flat lines of $\text{RL}^2$, PEARL in \textbf{humanoid-dir}, ProMP in \textbf{walker-rand-params}) via exploration or have unstable performance that drops after improvement (e.g., PEARL in \textbf{cheetah-vel} and \textbf{ant-goal}).

\subsection{Understanding CASTER's Behavior}\label{exp:und}

\subsubsection{Meta-overfitting}
As mentioned in \ref{sec:exp}, meta-overfitting arises due to train-test mismatch, an issue particular for 
meta-RL methods that incorporate off-policy learning. 
We compare the test-task performance obtained with different adaptation data distribution, i.e., the off-policy buffer data v.s. the on-policy exploration data, to see the performance gap when the data distribution shifts. We pick three environments that we find most prone to train-test mismatch, i.e., \textbf{cheetah-vel}, \textbf{ant-goal}, \textbf{point-robot}. Note that in the experiments, the number of transitions in the off-policy data equals the number of transitions in two episodes of the on-policy data.

Fig.~\ref{fig:meta-over} shows the results on the three environments\footnote{The \textbf{ant-goal} curve reproduced with PEARL's public repo has a discrepancy from the result reported in~\cite{rakelly2019efficient}, which we suspect to be the train-task performance with off-policy data. We recommend the reader to check with the publicly available code.}. CASTER takes a large leap 
towards bridging the gap between train and test sample distribution, reducing up to $75\%$ performance 
drop as compared to PEARL. This can be credited to better stochasity inherent in the explorer's episodes (Sec.~\ref{sec:vis-exp}) and better task inference of CASTER. The stochasity in the online rollouts brings its distribution closer to the off-policy data distribution, since at meta-train time data is randomly sampled from the buffer. Task inference that is aware of relations between samples better extracts the information in the few trajectories, which also narrow the gap by improving task inference.

\subsubsection{The Learned Exploration Strategy}\label{sec:vis-exp}
We investigate the exploration process by visualizing the histogram 
of the rewards in the collected trajectories. 
We take three consecutive trajectories rolled out by the explorer on 
two benchmarks : 1) the reward-varying environment \textbf{ant-goal}, 2) the transition-varying environment \textbf{walker-rand-params}. 
We compare with PEARL 
since it also resorts to posterior sampling for exploration~\cite{strens2000bayesian,osband2013more,osband2016deep}. 
\begin{figure}
  \centering
  \includegraphics[width=\columnwidth]{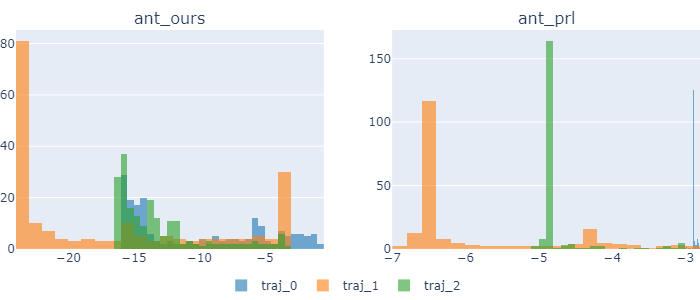}
  \caption{\small\textbf{Learned exploration behavior in \textbf{ant-goal}.} The reward histogram consists of 3 consecutive trajectories. 
  Left: CASTER and Right: PEARL. Different color denotes different trajectories.}\vspace{-3mm}
  \label{fig:vis-rew}
\end{figure}

In Fig.~\ref{fig:vis-rew}, it is shown that CASTER's exploration spans a large reward region of $[-25,0]$, while PEARL ranges from $-7$ to $-3$. This is consistent with our goal to increase the coverage of task experiences within few episodes, which potentially increases the chances to reach the goal and enables the task encoder to reason with both high rewards and low rewards, i.e., what region of state space might be the goal and might not. By contrast, PEARL tends to exploit the higher extreme values at the risk of following a wrong direction (e.g., in the second plot, PEARL proceeds to explore in a narrow region of relatively lower reward than the preceding episode).
\begin{figure}
  \centering
  \includegraphics[width=\columnwidth]{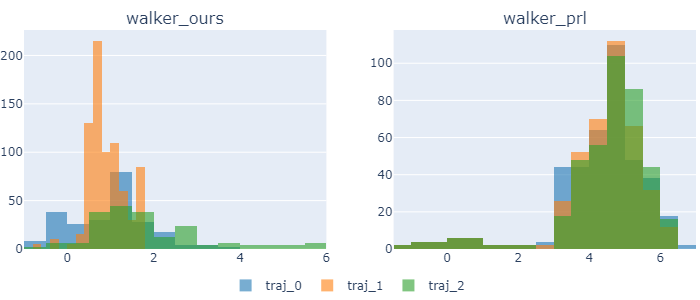}
  \caption{\small\textbf{Learned exploration behavior in \textbf{walker-rand-params}.} }
  \label{fig:vis-trans}
\end{figure}

In Fig.~\ref{fig:vis-trans}, the two methods are evaluated in the transition-varying environment \textbf{ant-goal} in which higher rewards do not necessarily carry the information of the system parameters. In this case, PEARL clings to higher reward regions as expected, while CASTER favors lower reward region. Supported by the results in Fig.~\ref{fig:sota},\ref{fig:test-eff}, PEARL's exploration is sub-optimal. CASTER performs better because the exploration is powered by information gain, and the explorer discovers task-discriminative experiences that happen to be around lower reward region.

\subsection{Ablations}\label{exp:abl}
In this section, we investigate our design choices of the exploration strategy and the task encoder via a set of ablative experiments on the \textbf{point-robot} environment.
We also provide the test-task performance 
with off-policy data (adaptation data from a buffer collected beforehand) to eliminate the effect of insufficient data source for task inference that induces train-test mismatch.
\subsubsection{The Task Encoder}\label{exp:inf}
\begin{figure}
  \centering
  \includegraphics[width=\columnwidth]{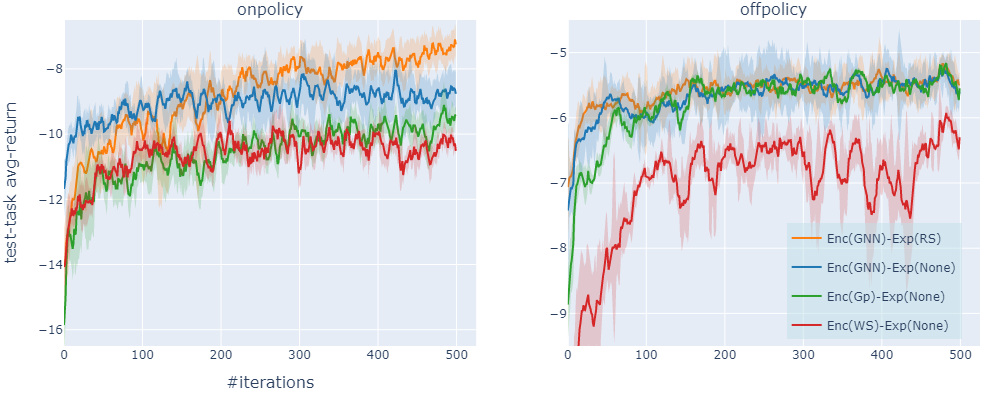}
  \caption{\small\textbf{Ablation study of the task encoder. We show the test-task performance with on-policy data (left) and off-policy data (right).}}
  \label{fig:inf}
  \vspace{-3mm}
\end{figure}

PEARL~\cite{rakelly2019efficient} builds its task encoder by stacking a Gaussian product (Gp) aggregator 
on top of an MLP, while we propose to combine self-attention 
with a learned weighted-sum aggregator for better relational modeling. 
We examine the following models: 1) \textbf{Enc(Gp)-Exp(None)} that uses the Gp task encoder, 3) \textbf{Enc(WS)-Exp(None)} that uses a learned weighted-sum aggregator without self-attention for graph message-passing,  3) \textbf{Enc(GNN)-Exp(None)} that uses the proposed GNN encoder, 4) and \textbf{Enc(GNN)-Exp(RS)} the proposed model. Note that for the first three baselines, the explorer is disabled to eliminate the impact of the exploration policy.

In Fig.~\ref{fig:inf}, all models tend to converge to similar asymptotic performance with off-policy data. However \textbf{Enc(WS)-Exp(None)} learns much slower than its counterparts, while \textbf{Enc(Gp)-Exp(None)} seems a reasonable design w.r.t. the off-policy performance. 
We conjecture that it can be hard for \textbf{Enc(Gp)-Exp(None)} to learn a weighting strategy as Gp, in which the weights are determined by the relative importance of a sample w.r.t. the whole pool of samples (${\sigma_i^2}/{\sum\sigma_j^2}$), and \textbf{Enc(Gp)-Exp(None)} has no access to other samples when computing the weight of each sample. By contrast, \textbf{Enc(GNN)-Exp(None)} provides more flexibility for the final weighted average pooling, by incorporating the interactions between samples. 
Such relational modeling enables it to extract more task statistics within the few episodes, superior to Gp in the on-policy performance.

\subsubsection{The Exploration Strategy}\label{exp:exp}
\begin{figure}
  \centering
  \includegraphics[width=\columnwidth]{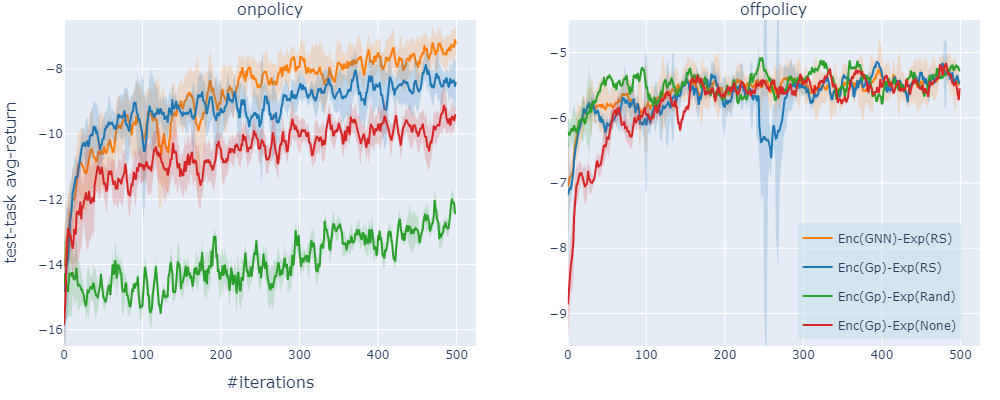}
  \caption{\small\textbf{Ablation study of exploration. We show the test-task performance with on-policy data (left) and off-policy data (right).}}
  \label{fig:exp}
  \vspace{-5mm}
\end{figure}

We aim to show the efficacy of the proposed reward shaping for task-exploration. The baseline models are: 1) \textbf{Enc(Gp)-Exp(None)} that uses no exploration policy, 
2) \textbf{Enc(Gp)-Exp(Rand)} that uses a (uniformly) random explorer, 3) \textbf{Enc(Gp)-Exp(RS)} that uses an explorer guided by the proposed reward shaping. Here the Gp task encoder is used by default.

As shown in Fig.~\ref{fig:exp}, all models perform equally in terms of 
asymptotic performance with off-policy data, 
since sufficient coverage of experiences are inherent in the data randomly sampled from buffers. This demonstrates the significance of improving coverage of task experiences for task inference.

For the on-policy performance, we can see \textbf{Enc(Gp)-Exp(Rand)} suffers from significant overfitting, 
as the coverage brought by the randomness doesn't suffice to benefit it within few episodes. \textbf{Enc(Gp)-Exp(RS)} performs much better than 
\textbf{Enc(Gp)-Exp(Rand)} because high rewards is an informative guidance in a reward-varying environment.
Our approach combines the merits of both broad coverage over task experiences and informative guidance regarding task relevance, and hence achieves the best performance.

\section{Conclusion}
We divide the problem of meta-RL into 
three sub-tasks, and take on probabilistic inference via variational 
EM learning. We thus present CASTER, a novel meta-RL method that learns dual agents with a task encoder for task inference. 
CASTER 
performs efficient task-exploration via a curiosity-driven 
exploration policy, from which the collected experiences are exploited by 
a context-aware task encoder. The encoder is equipped with the capacity for relational reasoning, with which the action policy adapts 
to complete the current task. Through extensive experiments, we show the superiority 
of CASTER over prior methods in sample efficiency, and empirically reveal that 
the learned exploration strategy efficiently acquires task-informative experiences with randomized behavior, which effectively helps mitigate meta-overfitting.
\section{Acknowledgement}
This work is
supported by Shanghai NSF Grant (No.~18ZR1425100) and NSFC
Grant (No.~61703195).

\clearpage
\bibliographystyle{ACM-Reference-Format}  
\bibliography{references}
\clearpage
\section{Supplemantary materials}
\subsection{Derivations of the Reward Function for Task Exploration}
Given a collected experience $c_t=(s_t,a_t,r_t,s_{t+1})$  obtained by executing $\xi(a_t|s_t,z_{t-1})$ in the environment. We use the following KL divergence to reflects the informativeness of $c_t$
\begin{equation}
    \begin{aligned}
        &\mathcal{D}_{KL}\left(q(Z|X=x_{t-1})\|q(Z|X=x_{t})\right) \\
        &= \mathcal{CE}\left(q(Z|X=x_{t-1})\| q(Z|X=x_{t})\right)-\mathcal{H}\left(q(Z|X=x_{t-1})\right)
      \end{aligned}
\end{equation}
where $x_{t}=x_{t-1}\cup c_{t}$, the entropy term $\mathcal{H}\left(q(Z|X=x_{t-1})\right)$ is constant w.r.t. the time step $t$.

The cross entropy term expands to the following expectation 
\begin{equation}
    \begin{aligned}
      &\mathcal{CE}\left(q(Z|X=x_{t-1})\| q(Z|X=x_{t})\right)\\
      &= -\int_z q(Z=z|X=x_{t-1})~\log~q(Z=z|X=x_t) dz \\
      &=\underset{\substack{z_{t-1}\sim q\left(Z|X=x_{t-1}\right)}}{\E}\left[-\log~q\left(Z=z_{t-1}|X=x_t\right) \right]
  \end{aligned}
  \end{equation}

  $q\left(Z=z_{t-1}|X=x_t\right)$ is found to relate with the Bellman Residual. By Bayes Rule
  \begin{equation}
    \begin{aligned}
        &P(z_{t-1}|x_{t-1},c_t)\cdot P(c_t)=P(c_t,z_{t-1}|x_{t-1})\\
        &=P(c_t|z_{t-1},x_{t-1})\cdot P(z_{t-1}|x_{t-1})
    \end{aligned}
  \end{equation}
we have 
\begin{equation}
    \begin{aligned}
      &P(z_{t-1}|x_t)\\
      &\propto P(c_t|z_{t-1},x_{t-1})\cdot P(z_{t-1}|x_{t-1})\\
      &={P(y_t|z_{t-1},s_t,a_t)\cdot P(a_t|s_t,z_{t-1})\cdot P(s_t)}
    \end{aligned}
  \end{equation}
  where we substitue $x_t$ for $x_{t-1}\cup c_t$. $c_t$ and $s_t$ is given when we compute the reward for exploration, so $P(c_t),P(s_t)$ will be ignored. $P(y_t|z_{t-1},s_t,a_t)$ relates with the Bellman Residual as discussed in Eq.~\eqref{eq:obj2}, $P(a_t|s_t,z_{t-1})\equiv\xi(a_t|s_t,z_{t-1})$. 

  As a result, the cross entropy has the following derivations:
  \begin{align}
    &\begin{aligned}
        &\underset{\substack{z_{t-1}\sim q\left(\cdot|x_{t-1}\right)}}{\E}\left[-\log~q\left(Z=z_{t-1}|X=x_t\right) \right]\\
    &\propto \underset{\substack{z_{t-1}\sim q\left(\cdot|x_{t-1}\right)}}{\E}\left[ -\log \left(P(y_t|z_{t-1},s_t,a_t)\cdot \xi(a_t|s_t,z_{t-1})\right)\right]\\
    &= \E_{z_{t-1}} \log \left(P(y_t|z_{t-1}^{(i)},s_t,a_t)\cdot \xi(a_t|s_t,z_{t-1}^{(i)})\right)\\
    &= \E_{z_{t-1}} \left[\lambda \left\|
    Q\left(s_t,a_t,z_{t-1}\right)-y_t
    \right\|^2 -\log \xi(a_t|s_t,z_{t-1})\right]\\
    &=\E_{z_{t-1}} \left[\lambda \left\|
    Q\left(s_t,a_t,z_{t-1}\right)-y_t
    \right\|^2 \right]+ \E_{z_{t-1}}\left[-\log \xi(a_t|s_t,z_{t-1})\right]\\
    &\defeq \hat{r}^\xi_t
    \end{aligned}\label{secondterm}
  \end{align}

To enforce stochasity, we formulate the following constrained RL objective based on $\hat{r}^\xi_t$:
\begin{equation}
    \begin{aligned}
      \E_{\rho(s,a,r)}\left[\sum_{i=0}^T \hat{r}_i^\xi \right]\quad \textrm{s.t.} \quad \mathcal{H}\left(\xi(\cdot|s_t,z_t)\right)\ge K
    \end{aligned}
  \end{equation}
  where $K$ is the policy entropy threshold. Notice that the policy entropy has the following form similar to the second term in Eq.~\eqref{secondterm}.
  \begin{equation}
    \mathcal{H}\left(\xi(\cdot|s_t,z_{t-1})\right) = \E_{a_t}\left[-\log \xi(a_t|s_t,z_{t-1})\right]
  \end{equation}

Since in implementation we optimize in the mini-batch manner with one Monte Carlo sample in each data point, we instead define the following reward function:
\begin{equation}
r^\xi_t = \lambda \left\|
Q\left(s_t,a_t,z_{t-1}\right)-y_t
\right\|^2  -\alpha~\log\xi(a_t|s_t,z_{t-1})
\end{equation}
which lead to the entropy-augmented soft Q-function~\cite{soft}. Via Lagrangian duality, the tuning of $\alpha$ amounts to the following minimization problem and can be achieved through SGD.
\begin{equation}
    \min_{\alpha} \E_{\xi^\ast} \left[-\alpha \log \xi^\ast (a_t|s_t,z_{t-1};\alpha_0)-\alpha K\right]
\end{equation}
where $\xi^\ast$ is the optimized policy with the dual variable fixed at $\alpha_0$.

\end{document}